\pdfoutput=1

\documentclass[11pt]{article}

\usepackage{EMNLP2023}

\usepackage{times}
\usepackage{latexsym}
\usepackage{amsmath}
\usepackage{multirow}
\usepackage{subfig}
\usepackage{xspace}
\usepackage{tikz}
\usepackage{pgf-pie}

\usepackage[T1]{fontenc}

\usepackage[utf8]{inputenc}

\usepackage{microtype}

\usepackage{inconsolata}

\usepackage{array}
\usepackage{graphicx}
\usepackage{makecell}
\usepackage{float}
\usepackage{enumitem}

\newcommand{\our}{\textsc{StoC-ToT}\xspace}

\title{\our: Stochastic Tree-of-Thought with Constrained Decoding for Complex Reasoning in Multi-Hop Question Answering}

\author{Zhenyu Bi \\
  Virginia Tech \\
  \texttt{zhenyub@vt.edu} \\\And
  Daniel Hajialigol \\
  Virginia Tech \\
  \texttt{danielhajialigol@vt.edu} \\\And
  Zhongkai Sun \\
  Amazon Alexa AI \\
  \texttt{zhongkas@amazon.com} \\\AND
  Jie Hao \\
  Amazon Alexa AI \\
  \texttt{jieha@amazon.com} \\\And
  Xuan Wang \\
  Virginia Tech \\
  \texttt{xuanw@vt.edu} \\}

\begin{document}
\maketitle
\begin{abstract}
Multi-hop question answering (MHQA) requires a model to retrieve and integrate information from multiple passages to answer a complex question. Recent systems leverage the power of large language models and integrate evidence retrieval with reasoning prompts (e.g., chain-of-thought reasoning) for the MHQA task. However, the complexities in the question types (bridge v.s. comparison questions) and the reasoning types (sequential v.s. parallel reasonings) require more novel and fine-grained prompting methods to enhance the performance of MHQA under the zero-shot setting.
In this paper, we propose \our, a stochastic tree-of-thought reasoning prompting method with constrained decoding for MHQA and conduct a detailed comparison with other reasoning prompts on different question types and reasoning types. Specifically, we construct a tree-like reasoning structure by prompting the model to break down the original question into smaller sub-questions to form different reasoning paths. In addition, we prompt the model to provide a probability estimation for each reasoning path at each reasoning step. At answer time, we conduct constrained decoding on the model to generate more grounded answers and reduce hallucination. Experiments comparing \our with on two MHQA datasets and five large language models showed that \our outperforms other reasoning prompts by a significant margin. 
\end{abstract}

\section{Introduction}
\begin{figure}
\includegraphics[width=6.5cm, height=6.5cm]{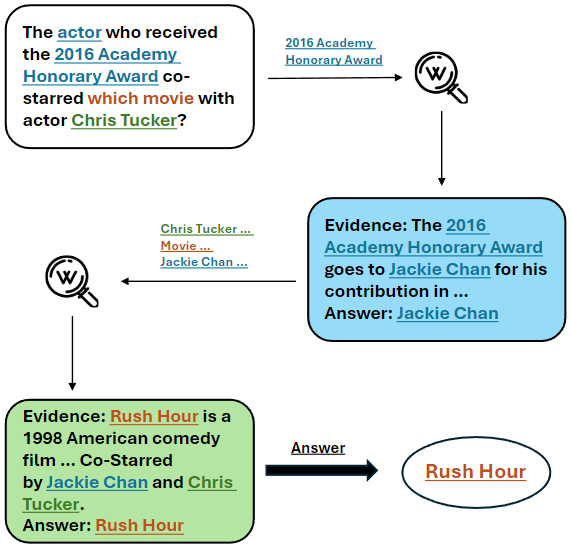}
\centering
\caption {An example of the MHQA question. This question has two hops that require the model to reason about before answering the final question.}
\label {Fig1}
\end{figure}

Question answering (QA) is a fundamental task in natural language processing (NLP) that involves designing systems capable of understanding human language questions and providing accurate and relevant answers. With the recent advancement of large language models (LLMs) that demonstrated superior reasoning ability \cite{gpt3}, researchers have been focusing more on complex QA tasks, such as multi-hop question answering (MHQA). MHQA is more challenging as it requires models to understand complicated questions, perform multiple reasoning steps, and gather evidence across documents. Figure \ref{Fig1} shows an example of a two-hop MHQA question. To answer that question in Figure \ref{Fig1}, the QA model needs to first figure out who is the actor that received the 2016 Academy Honorary Award. Then based on the answer to the previous question, the QA model needs to further answer a second question about which movie the actor co-starred with Chris Tucker.

State-of-the-art methods for MHQA are fully-supervised methods that often follow a retrieve-and-read framework, including a passage retrieving module that gathers relative evidence from documents and a reading comprehension module to reason about the evidence \cite{zhu2021, Li2022FromEH}. Other methods include beam-search \cite{beam} and label-smoothing \cite{YinWHWYZCHQ23}. However, these methods often require extensive pre-training or fine-tuning and do not generalize well to other datasets.

Large language models (LLMs), on the other hand, show remarkable reasoning ability and rich knowledge of general-domain questions. Many LLMs can answer simple and straightforward questions that do not require complex reasoning without any supervision involved but often fail to deal with complex questions requiring multiple reasoning steps. To tackle the problem, researchers have developed many prompting techniques to improve LLM's reasoning ability, such as chain-of-thought (CoT) \cite{CotWei}, self-consistency CoT (Sc-CoT) \cite{sccot}, and tree-of-thought (ToT) prompting \cite{TOT23}. 

CoT has been shown effective across tasks requiring extensive, step-by-step reasoning, such as math calculation and reading comprehension. However, there could be various possible reasoning paths for many complex multi-hop questions, and CoT models cannot "turn back" when they have made a mistake along their reasoning paths. Sc-CoT further improves on CoT by proposing different chains of thought, thus expanding the reasoning space. However, there is no local reasoning expansion within each chain, and the "majority voting" strategy often fails in open-domain tasks where the output space is unlimited. ToT, designed to maintain different reasoning paths along its reasoning process, is more suitable for dealing with complex question types. However, the intermediate reasoning steps in NLP generation tasks are much less constrained and require more than a simple rule-based evaluation. The complexities in the question types (bridge v.s. comparison questions in Table \ref{table:1}), as well as the reasoning types (sequential v.s. parallel reasonings in Table \ref{table:2}), require more novel and fine-grained prompting methods to enhance the reasoning ability of LLMs. 

\begin{figure*}[t]
\includegraphics[width=16cm, height=8.5cm]{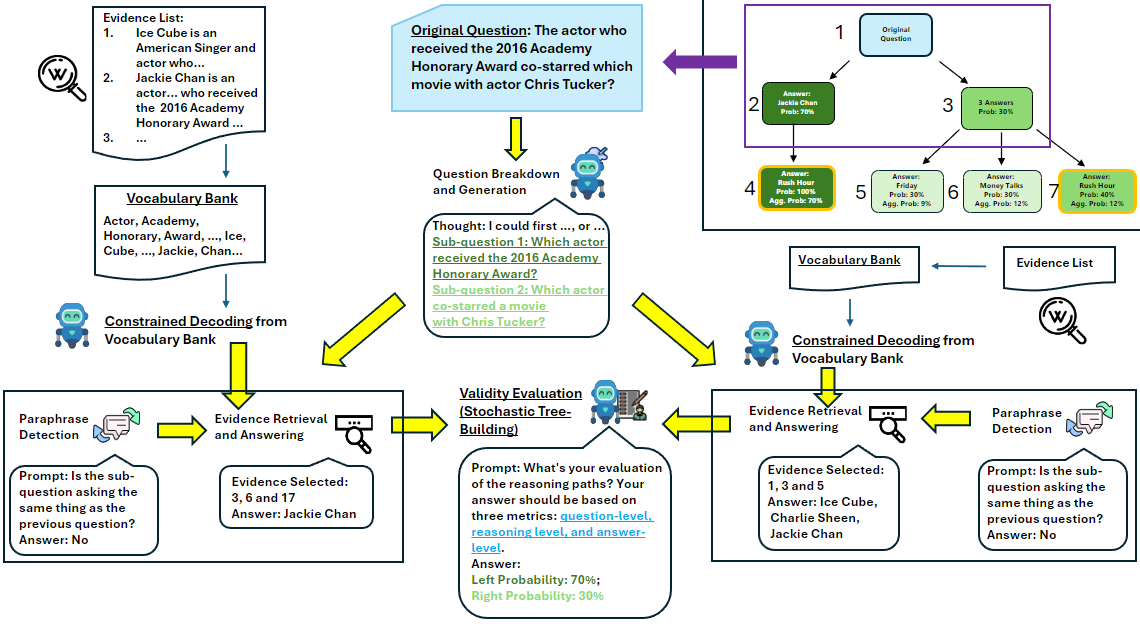}
\centering
\caption {Overview of our framework, with the example in Figure 1. The top-right Corner shows the overall structure of the constructed tree, with each node's label on the left. \textbf{Darker green} in the nodes means a \textbf{higher evaluated probability of the reasoning path}. The original Question is colored in blue. We chose the first round of our tree-building process as an example in the purple block. }
\label {Fig2}
\end{figure*}
To tackle the challenges and design a more reliable reasoning method for open-domain NLP tasks, we propose \our, a stochastic ToT-based framework that instructs the model to generate different reasoning paths from the same question and assign probability scores to reasoning paths to effectively avoid reasoning dead-ends. To the best of our knowledge, our work is the first to adapt the tree-of-thought reasoning prompting to natural language tasks that require complex reasoning, such as MHQA. We provide an example overview of our framework in Figure \ref{Fig2}.
Specifically, we construct a tree-like reasoning structure by prompting the model to break down the original question into smaller sub-questions to form different reasoning paths. We evaluate the validity of each reasoning path on three levels of aspects and arrive at a model-given probability score. At answer time, we innovatively propose to use constrained decoding in the answering process to reduce hallucination by forcing the model to generate grounded answers from evidence and letting models give concise and exact answers. Ultimately, we arrive at the best answer by choosing the path with the highest aggregated probability score. Experiments on two benchmarking MHQA datasets demonstrate that \our significantly improves the reasoning ability of LLMs in complex reasoning scenarios, especially with GPT-4, improving Exact Match accuracy by 7\%, and F1 score by 7.8 points on the HotpotQA dataset over the original tree-of-thought prompting. 
Our contributions are as follows:
\begin{itemize}[leftmargin=*]
    \item We propose \our, which constructs a stochastic reasoning tree in complex reasoning scenarios. We introduce stochastic estimations on different reasoning paths, which helps the model have a more reliable reasoning process than previous reasoning prompting methods.
    \item We innovatively propose to use constrained decoding in the answering process. This step reduces model hallucination by forcing the model to generate grounded answers from evidence and letting models give concise and exact answers.
    \item We evaluate the effectiveness of \our by conducting experiments on two MHQA datasets. We observe substantial improvements over other reasoning prompting methods, with \our surpassing all other selected reasoning prompting baselines on 5 tested models.
\end{itemize}


\section{Related Work}
\paragraph{Multi-Hop Question Answering}
Multi-hop Question Answering (MHQA) is a challenging task requiring models to reason over different evidence across documents to answer a complex multi-hop question. Many high-quality MHQA datasets have been developed, including HotpotQA \cite{yang2018hotpotqa}, WikiHop \cite{WelblSR18}, MuSiQue \cite{TrivediBKS22}, and others. Among these, HotpotQA is the task's most representative and widely used dataset.
Previous state-of-the-art MHQA models often follow a two-stage pipeline: a retriever that extracts evidence from the documents, and a reader that reasons about the evidence to arrive at an answer \cite{zhu2021, Li2022FromEH}. Other methods include beam-search \cite{beam} and label-smoothing \cite{YinWHWYZCHQ23}. Some LLM-based frameworks \cite{react,critic} were also evaluated on the task of MHQA, but their performance fell short compared with supervised methods.

\paragraph{Reasoning Prompting of LLMs}
Various prompt engineering methods have been developed \cite{CotWei,sccot,TOT23,Got24,Aot24,Pot23}, aiming to improve large language models' reasoning ability across various tasks and domains. 
Chain-of-thought (CoT) prompting \cite{CotWei} prompts the large language models (LLMs) to divide their reasoning process into smaller steps when solving a question, forming a chain of thoughts. Chain-of-thought self-consistency prompting \cite{sccot} improves on the CoT method by proposing different reasoning chains and ensembles on the final result.
Tree-of-thought (ToT) prompting method \cite{TOT23} actively maintains a tree of thoughts, where each thought is a coherent language sequence that serves as an intermediate step toward problem-solving. Graph-of-thought \cite{Got24} further improves ToT by constructing a Directed Graph instead of a tree. LLMs can loop over a thought to refine it and aggregate thoughts or chains. 

\paragraph{Constrained Decoding} Constrained decoding is the technique that asks the models to generate outputs following a given set of rules. The most common way of conducting constrained generation uses beam search \cite{beamori} in decoding time. Before the LLM era, works on constrained decoding focused on task-specific sequence-to-sequence models that span across many fields, such as machine translation \cite{cdbeam,cdbeam2}, named entity recognition \cite{cdbeamNER}, and dialogue generation \cite{cdDialogue}. Recently, Microsoft introduced Guidance \footnote{https://github.com/guidance-ai/guidance}, which allows users of various large language models to control their outputs given a human-defined vocabulary or rules.

\section{Method}

\subsection{Task Formation}
Given a multi-hop question $Q$ and background corpus of evidence $P$, the goal of our framework is to output the answer $A$ to question $Q$, drawing its reasoning with the support of multiple evidence passages $p_1, p_2, ...$ retrieved from corpus $P$.


\subsection{\our Framework}
For each of the questions $Q$, multiple reasoning lines and, thus, multiple ways of breaking down the question could exist. However, not every reasoning line would lead us to the right answer, and they take us to dead ends. To avoid such reasoning dead-ends, we build a stochastic reasoning tree to represent the possible reasoning lines and the probability of each reasoning line taking us to the right answer. We achieve this by proposing a self-interactive framework that automatically builds the reasoning tree given a multi-hop question. Figure \ref{Fig2} shows our framework with an example question.

In our reasoning process, we first prompt the model to propose different possible sub-questions to solve at each reasoning step. Each sub-question corresponds to one possible reasoning path and is presented as a node in the tree. We then ask the model to answer the generated sub-questions. To prevent hallucination and make the model more focused on the given question and evidence, we build a vocabulary bank using words from the evidence list and the original question and instruct the model to do constrained decoding from the vocabulary bank when generating its answers. After answering every sub-question generated from the same question in the previous reasoning level, we prompt the model to evaluate each reasoning path and estimate how likely the reasoning path would lead us to the right answer. This probability estimation would be assigned to the corresponding node in the tree. After the reasoning process finishes, each reasoning path would have an aggregated probability calculated from nodes along the path.

Formally, given a question $Q$, we instruct the model to generate sub-questions $q_1, q_2, ..., q_n$, and build a tree structure with the original question $Q$ as the root node and each question $q_i$ as subsequent nodes. The tree would expand as each sub-question $q_i$ has its sub-question $q_j$, and the reasoning paths are thus represented as branches in the tree structure. From the original question $Q$ and the evidence list $E = e_1, e_2, ..., e_n$, we build a vocabulary bank $V=[w_1, w_2, ..., w_n],w_i\in{Q},w_j\in{E}$. We then prompt the model to generate their answer $a_1, a_2, ..., a_n$ using only $w_i\in{V}$. We describe the details of our framework below.


\paragraph{Example-Based Sub-Question Generation}
Our framework starts with the sub-question generation module, which generates sub-questions $q_1, q_2, ..., q_n$ using the question $Q_g$ from the previous reasoning level. The sub-questions are generated based on both the model's reasoning ability and the model's semantic understanding of the question $Q_g$. An example is given in Figure \ref{Fig2}, where the sub-questions from nodes 2 and 3 were generated using the question from node 1. However, we cannot guarantee that each sub-question asked is a good sub-question, and sometimes, the generated sub-question merely repeats the previous question. We introduce the paraphrase detection module and pass on the generated sub-questions to reduce redundancy and improve question quality.

\paragraph{Paraphrase Detection}
Answering repetitive questions often leads to low-quality answers and time-consuming steps. Following the sub-question generation module, we introduce the paraphrase detection module to reduce redundancy and improve question quality. In this module, we prompt the model and ask it to distinguish informative questions from questions that merely repeat what is already stated at the previous reasoning level. If a sub-question is a paraphrase, we instruct the model to stop generating sub-questions from the current question. In other words, we prune the low-quality sub-branch of the tree that could otherwise be generated. By pruning these branches, we effectively improve the efficiency of our framework. 

\paragraph{Evidence Retrieval and Answering}
We then move on to answering the question after our paraphrase detection module. Our evidence retrieval and answering module focuses on retrieving evidence and generating answers to the given sub-question. We also pass in the full evidence list provided and prompt the model to give out an answer to the given sub-question. The evidence retrieval and answering module selects relative evidence from an evidence pool for each sub-question and uses words only from the vocabulary bank to generate its final answer. We will discuss details of constrained decoding in Section \ref{sec:constrained}. The generated sub-answer and the answered sub-question are then passed on to the sub-question generation module at the next level to continue the reasoning process.

\paragraph{Validity Estimation}
Not each sub-question asked is a good sub-question, and not each reasoning path is reasonable. After every sub-question $q_i$ generated from the same question $Q_g$ has been answered, we prompt the model to provide a probability estimation $p_i$ for each $(q_i, a_i)$ pair. This probability is the model's evaluation of going down the correct reasoning path. Specifically, this probability is obtained by prompting the model to consider the following three aspects:
\begin{itemize}[leftmargin=*]
    \item Question Level: Is the question semantically clear and answerable? 
    \item Reasoning Level: Is the reasoning line coherent when considering previous levels? 
    \item Answer Level: Does the evidence fully support the answer to the question? 
\end{itemize}
As shown in Figure \ref{Fig2}, we conduct validity estimation for sub-questions and sub-answers in nodes 2 and 3 since the sub-questions were generated from the same question in node 1. 

At the leaf node of our tree, we would have a final question $q_f$. along with a final answer $A$ to the original question $Q$, and also an aggregated probability $p_{final}=\prod_{i}p_{i}$, with each $p_i$ being the probability of the nodes along the reasoning path. We assign $p_{final}$ to the leaf node, representing the aggregated probability of answer $A$ being the correct answer to $Q$.

\subsection{Constrained Decoding} \label{sec:constrained}
One challenge for generative LLMs in the task of question answering is hallucination. LLMs often fail to pay attention to the golden evidence and hallucinate their own reference even when large amounts of evidence exist. To alleviate the problem of LLM halluscination during evidence selection and answer generation, we innovatively propose to use constrained decoding in the answering process to reduce hallucination by forcing the model to generate grounded answers from evidence and let models give concise and exact answers. As shown in Figure \ref{Fig2}, we conduct constrained decoding by asking the model to generate words from the vocabulary bank, consisting of words taken only from the original question and the evidence list provided. More formally, we construct a vocabulary bank $V={w_1,w_2,...,w_i}$ from all words in the provided evidence sentences. We conduct a simple filtering by removing common English stop words. We then instruct the model's evidence retrieval and answering module to construct its answers using words only from the given vocabulary $V$.

\paragraph{Code-based Constrained Decoding} For open-source LLMs (e.g., Llama), we build our logit processor at the decoding time. Specifically, for every word $w_j\notin{V}$, we manually set the score to negative infinity to prevent the model from generating them. Thus, every answer generated will only use words from the evidence list.

\paragraph{Prompt-based Constrained Decoding} For closed-source LLMs (e.g., GPT models), since we do not have access to their decoding function, we had to instruct the GPT models using prompts to do constrained decoding. We provide our prompt template used in Appendix \ref{sec:appendixA}.

\section{Experimental Setup}
\begin{table*}[ht]
\caption {Performance comparison of \our and baseline methods on the HotpotQA dataset.}
\centering
\resizebox{\textwidth}{!}{
\begin{tabular}{c|cc|cc|cc|cc|cc}
\hline
\multirow{2}{*}{Prompting Method} & \multicolumn{2}{c|}{GPT3.5} & \multicolumn{2}{c|}{GPT4} & \multicolumn{2}{c|}{LLaMa2(13B)} & \multicolumn{2}{c|}{LLaMa2(70B)} & \multicolumn{2}{c}{LLaMa3(8B)}\\
\cline{2-11} & EM & F1 & EM & F1 & EM & F1 & EM & F1 & EM & F1\\
\hline
Zero-Shot Vanilla & 34.0 & 45.0 & 51.0 & 65.0 & 25.5 & 36.5 & 30.5 & 41.0 & 27.5 & 40.7\\
Chain-of-Thought & 35.5 & 47.3 & 52.0 & 66.8 & 30.5 & 42.5 & 33.5 & 45.0 & 32.5 & 44.6\\
Tree-of-Thought & 36.5 & 49.5 & 55.0 & 68.5 & 29.5 & 41.3 & 35.5 & 47.3 & 30.5 & 37.5\\ \hline
\textbf{\our} & \textbf{45.5} & \textbf{56.2} & \textbf{62.0} & \textbf{76.3} & \textbf{31.0} & \textbf{43.0} & \textbf{43.0} & \textbf{56.3} & \textbf{33.0} & \textbf{44.5}\\ 
\textbf{w/o constrained decoding} & 40.5 & 53.5 & 59.5 & 73.0 & 31.0 & 43.0 & 40.5 & 53.5 & 32.0 & 44.3\\ \hline
\end{tabular}
}
\label{table:1}
\end{table*}

\begin{table*}
\caption {Performance comparison of \our and baseline methods on the MusiQue dataset.}
\centering
\begin{tabular}{c|cc|cc|cc|cc}
\hline
\multirow{2}{*}{Prompting Method} & \multicolumn{2}{c|}{GPT3.5} & \multicolumn{2}{c|}{GPT4} & \multicolumn{2}{c|}{LLaMa2(13B)} & \multicolumn{2}{c}{LLaMa3(8B)}\\
\cline{2-9} & EM & F1 & EM & F1 & EM & F1 & EM & F1\\
\hline
Zero-Shot Vanilla & 17.0 & 28.8 & 31.5 & 41.2 & 9.5 & 16.0 & 12.0 & 19.2 \\
Chain-of-Thought & 18.0 & 29.7 & 32.5 & 44.2 & 11.0 & 17.5 & 12.5 & 21.6 \\
Tree-of-Thought & 20.5 & 32.0 & 35.0 & 47.3 & 11.0 & 17.2 & 12.0 & 20.6 \\ \hline
\textbf{\our} & \textbf{26.5} & \textbf{38.0} & \textbf{42.0} & \textbf{55.3} & \textbf{11.5} & \textbf{18.0} & \textbf{14.5} & \textbf{22.0} \\
\textbf{w/o constrained decoding} & 24.0 & 35.5 & 38.5 & 51.0 & 11.5 & 18.0 & 14.0 & 22.0 \\ \hline
\end{tabular}
\label{table:2}
\end{table*}

\paragraph{Dataset} We compare \our with baseline methods on the HotpotQA dataset \cite{yang2018hotpotqa} and the MuSiQue dataset \cite{TrivediBKS22}, both of which are widely used MHQA datasets across state-of-the-art MHQA baselines. The experiments are conducted under the distractor setting, where we provide the model with an evidence pool containing both golden and irrelevant evidence. The model needs to find the golden evidence to answer the question correctly. We randomly selected 200 examples from each dataset as our evaluation set.

\paragraph{Baselines} We included three baselines:
\begin{itemize}[leftmargin=*]
    \item Vanilla Prompting with no examples provided. We only provide the model with questions and evidence and instruct it to output the answer.
    \item Chain-of-Thought (CoT) prompting \cite{CotWei} with a standard input-output (IO) prompt. We design the prompt with one in-context example, which presents the whole reasoning chain, including all intermediate steps.
    \item Tree-of-Thought prompting \cite{TOT23} with slight modifications to adapt to the MHQA task. We largely followed the original framework and used majority voting on the reasoning lines to decide the final answer. 
\end{itemize}
We recognize that there are LLM-based retrieval augmented generation frameworks \cite{react, critic} that were also evaluated on HotpotQA. However, we excluded them from our baselines as they used outside knowledge bases, which are under a different testing scenario.

\subsection{Implementation} We experiment with the baselines and our model utilizing five LLMs: GPT-3.5-turbo \cite{gpt3} and GPT-4\cite{openai2024gpt4} from OpenAI, LLaMa 2-13B \cite{llama2}, LLaMa 2-70B, and LLaMa 3-8B from MetaAI. Due to the lengthy running time, LLaMa 2-70B was not tested on the MusiQue dataset. For all models, We set the temperature to 0.5, $top_k$ to 1.0, and maximum number of iterations to 5.

\subsection{Evaluation Metric} Following the metrics in \cite{yang2018hotpotqa}, we use Exact Match and F1 score as two evaluation metric. For an answer $a$ given by our framework, the Exact Match score equals 1 if the answer span matches the golden answer exactly and 0 otherwise. The F1 metric measures the average overlap between the prediction and ground truth answers. 

\section{Results}
\subsection{Overall Results}
We compare \our with LLM baselines on the HotpotQA dataset and the MusiQue dataset and present our results in Tables \ref{table:1} and \ref{table:2}. The backbone LLMs in our experiments include GPT3.5, GPT4, Llama2-13B, Llama2-70B, and Llama3-8B. Due to time constraints, we only tested with Llama2-70B on the HotpotQA dataset. On the HotpotQA dataset, \our attains an on-average increase in performance of over 6 \% compared with vanilla prompting on GPT models, and the improvement goes up to 11\% when we further implement \our with constrained decoding. On the more challenging MusiQue dataset, we still see an increase in performance of \our compared with the other baselines, most notably on GPT4, where we observe an 11.5\% EM improvement (from 31.50 to 42.0). 

\paragraph{Comparison with Tree-of-Thought} \our surpasses the original Tree-of-Thought prompting by 7\% with the GPT4 model on both tested datasets. For LLMs with inferior reasoning ability, such as LLaMa2-8B, we still observe a performance improvement, even on the harder MusiQue dataset. These results suggest that \our is more effective at forming and selecting reliable reasoning paths under complex reasoning scenarios.

\paragraph{Constrained Decoding} Even though the LLM's reasoning ability can be improved by reasoning prompting, such techniques have little help in preventing hallucination. However, \our implements constrained decoding, which makes the model much more grounded to evidence when answering the question, effectively addressing hallucination issues and improving the overall performance of our framework.

\subsection{Ablation Study}

\paragraph{Sensitivity to Demonstration Question Type}
\begin{table*}[t]
\caption {Performance of \our with different prompt types on the HotpotQA dataset in terms of EM score. ``Com" represents comparison questions, and ``Bri" represents bridge questions.}
\centering
\resizebox{\textwidth}{!}{
\begin{tabular}{c|c|c|c|c|c|c|c|c|c|c}
\hline
\textbf{Model Variant} & \multicolumn{2}{c|}{\textbf{GPT3.5}} & \multicolumn{2}{c|}{\textbf{GPT4}} & \multicolumn{2}{c|}{\textbf{LLaMa2(13B)}} & \multicolumn{2}{c|}{\textbf{LLaMa2(70B)}} & \multicolumn{2}{c}{\textbf{LLaMa3(8B)}} \\
\hline
\textbf{Prompt/Question Type} & \textbf{Com} & \textbf{Bri} & \textbf{Com} & \textbf{Bri} & \textbf{Com} & \textbf{Bri} & \textbf{Com} & \textbf{Bri} & \textbf{Com} & \textbf{Bri}\\
\hline
Prompt: Comparison & 58.8 & 41.0 & 76.5 & 57.2 & 38.2 & 31.9 & 58.8 & 41.0 & 44.1 & 33.7 \\ \hline
Prompt: Bridge & 55.9 & 43.4 & 73.5 & 59.0 & 35.3 & 32.5 & 55.9 & 42.2 & 41.2 & 34.9\\
\hline
\end{tabular}
}
\label{table:3}
\end{table*}

We study the effect on \our performance when different types of demonstration questions are provided in the prompt template. The HotPotQA dataset specified two types of questions. The "Bridge" question contains a "bridge entity” that connects the question and the final answer. In contrast, the "Comparison" question requires the model to compare two entities of the same type. Of the 200 questions in our evaluation set, 34 are comparison questions, and 166 are bridge questions. Examples of bridge and comparison questions are in Table \ref{table:4}.

We examined \our performance under the two different question types, each with a different prompt template: one containing only a comparison question as an example and the other containing only a bridge question as an example. We provide the content of our templates in Appendix \ref{sec:appendixA}. Results are shown in Table \ref{table:3}. We observe that the difference in prompt templates influences the performance of our framework under different question types by a small margin. The comparison questions are generally easier to solve, and \our performs better on comparison questions than on bridge questions. \our will handle comparison questions better if the prompt template contains comparison questions and vice versa. 

\paragraph{Question and Reasoning Types}
\begin{figure*}[t]
    \centering
    \subfloat[\centering Question Type]{{\includegraphics[width=7cm]{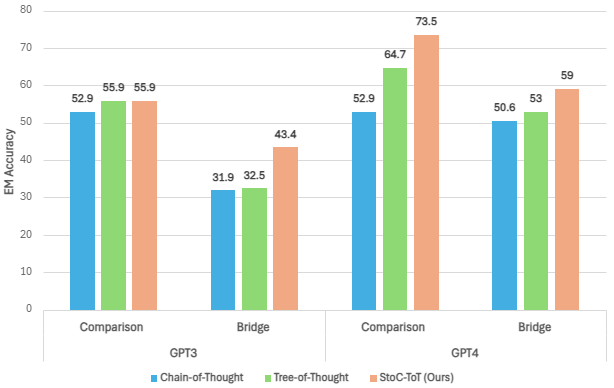} }}%
    \label{fig3a}
    \qquad
    \subfloat[\centering Reasoning Type]{{\includegraphics[width=7cm]{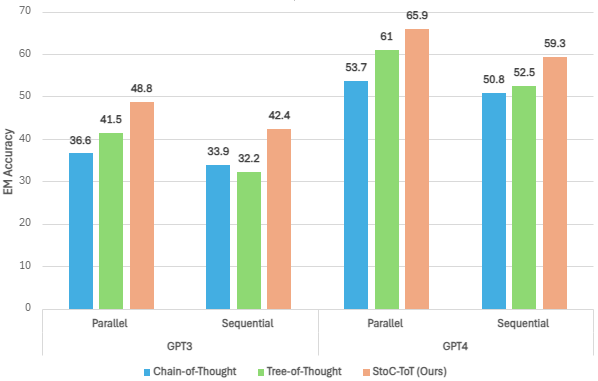} }}%
    \label{fig3b}
    \caption{Performace comparison of Chain-of-Thought, Tree-of-Thought, and {\our} on questions of different question types (Left) and reasoning types (Right). Experiments were done on the HotpotQA dataset.}%
    \label{fig3}%
\end{figure*}

\begin{table}[t]
\caption {Question Type Examples. On the left side, the bridging entity is highlighted in red, and the final question is highlighted in orange. On the right side, entities that are being compared are highlighted in blue.}
\centering
\resizebox{.5\textwidth}{!}{
\begin{tabular}{m{4cm}|m{4cm}}
\hline
\textbf{Bridge Question} & \textbf{Comparison Question}\\
\hline
{\textcolor{orange}{What distinction} is held by \textcolor{red}{the former NBA player} who was a member of the Charlotte Hornets during their 1992-93 season and was head coach for the WNBA team Charlotte Sting?} & Were \textcolor{blue}{Scott Derrickson} and \textcolor{blue}{Ed Wood} of the same nationality? \\ \hline
\end{tabular}
}
\label{table:4}
\end{table}

\begin{table}[t]
\caption {Reasoning Type Examples. On the left side, the entity in red needs to be found before solving the question in orange. On the right side, questions with parallel reasoning contain parts (highlighted in blue) that can be solved in arbitrary order.}
\centering
\resizebox{.5\textwidth}{!}{
\begin{tabular}{m{4cm}|m{4cm}}
\hline
\textbf{Sequential Reasoning} & \textbf{Parallel Reasoning}\\
\hline
The \textcolor{red}{football manager} who recruited David Beckham managed Manchester United during \textcolor{orange}{what timeframe}? & What distinction is held by the former NBA player who was a \textcolor{blue}{member} of the Charlotte Hornets during their 1992-93 season and was \textcolor{blue}{head coach} for the WNBA team Charlotte Sting? \\ \hline
\end{tabular}
}
\label{table:5}
\end{table}

We examine \our, Tree-of-Thought prompting, and Chain-of-Thought prompting by comparing their performance under different question-type settings. Detailed results are shown in Figure \ref{fig3}(a). \our performs better at both Bridge Questions and Sequential Questions, suggesting that \our can avoid reasoning dead-ends and is better at forming intermediate reasoning lines. 

We also conduct an in-depth analysis of the reasoning types in the existing MHQA datasets by randomly selecting 100 questions from our testing set. The questions are roughly divided into two categories: 1) tree-like parallel reasoning and 2) chain-like sequential reasoning. Questions with parallel reasoning contain two or more reasoning paths that can be solved arbitrarily. Questions with sequential reasoning follow a strict reasoning chain, and all the sub-questions must be solved to form the correct reasoning process. All comparison questions are parallel reasoning, but some bridge questions contain parallel reasoning. Examples of sequential and parallel reasoning questions are in Table \ref{table:5}. Out of the selected 100 questions, 59 questions were Sequential and 41 questions were Parallel. Results are shown in Figure \ref{fig3}(b). \our performs better on both reasoning types, especially on questions containing parallel reasoning. This suggests that \our's stochastic way of forming the tree is very effective when solving questions containing multiple reasoning paths.

\paragraph{Performance and Hops} 
\begin{figure}[t]
\resizebox{.5\textwidth}{!}{
\includegraphics{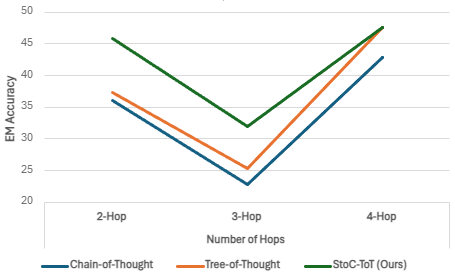}
}
\centering
\caption {Performance comparison of CoT, ToT, and \our on different number of hops in the question. Experiments done in the MusiQue dataset.}
\label{fig4}
\end{figure}

As the number of hops increases in a question, the reasoning line gets more complex and varied. Figure \ref{fig4} shows the performances of different prompting techniques on questions in the MusiQue dataset with different numbers of hops. \our performs best in all categories, demonstrating our framework's superior ability to deal with complex reasoning scenarios. This ablation study was conducted only on GPT4, as other models performed poorly on 3-hop and 4-hop scenarios, regardless of the reasoning prompting technique used.

\paragraph{Error Analysis} 
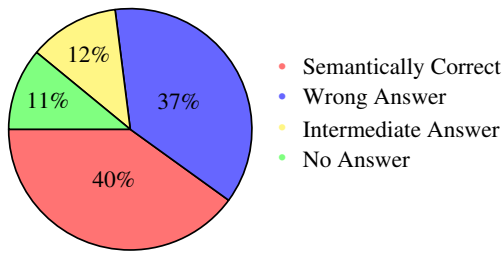
\begin{figure}[t]
\centering
\scalebox{0.8}{
\begin{tikzpicture}
    \pie[rotate=180, color={red!55, blue!60, yellow!60, green!50}, radius=2]
    {40/, 37/, 12/, 11/}
    \begin{scope}[shift={(2.5cm,1cm)}]
            \node[red!55, label=right:Semantically Correct] at (0,0) {\textbullet};
            \node[blue!60, label=right:Wrong Answer] at (0,-0.5) {\textbullet};
            \node[yellow!60, label=right:Intermediate Answer] at (0,-1) {\textbullet};
            \node[green!50, label=right:No Answer] at (0,-1.5) {\textbullet};
    \end{scope}
\end{tikzpicture}
}
\caption{Ratio of different categories in error cases, on the HotpotQA dataset.}
\label{fig5}
\end{figure}
We conduct a detailed analysis of the errors made by our framework on GPT3 and GPT4, and present our results in Figure \ref{fig5}. We categorize the errors into four types: (1) \textbf{No Answer}: our framework did not come up with an answer for the question due to not finishing the reasoning process; (2) \textbf{Intermediate Answer}: our framework came up with an answer for one of the intermediate hops instead of for the final question; (3) \textbf{Wrong Answer}: our framework came up with an answer that is neither the final answer nor one of the intermediate answers; (4) \textbf{Semantically Correct}: our framework came up with the right answer, but did not have an exact match with the final answer. Appendix \ref{sec:appendixB} shows examples of each error category. Large amounts of error cases were correct answers with extra wording or hallucination errors, signaling potential improvements over our constrained decoding scheme. Reasoning process errors, including no answer and intermediate answer, make up only 25\% of the total error cases. This result shows that our framework is capable of building a robust reasoning process for complex questions.

\section{Conclusion}
This paper proposes \our, a stochastic tree-of-thought reasoning framework with constrained generation for multi-hop question answering. \our is specialized in dealing with complex reasoning scenarios in natural language tasks. Experiments on two benchmark datasets show that our framework outperforms previous reasoning prompting techniques with multiple Large Language Models. Detailed analysis shows that our framework is capable of building a robust reasoning process given different types of questions. Further research can aim to enhance the reliability of our framework by proposing better validity evaluation schemes and more effective methods for improving groundedness and preventing hallucination.

\clearpage
\section*{Limitations}
Our framework relies on initiating multiple model instances and requires multiple prompts per round. The repetitive callings impose heavy time costs for our framework, even after implementing our paraphrase module. Another limitation comes from how we generated sub-questions. Currently, we directly prompt the model to generate sub-questions. A more complex standard can be used to increase the quality of the sub-questions generated. Also, more extensive experiments should be provided, including experimenting on other different datasets and case studies.
\section*{Ethics Statement}
This research adhered to the ethical standards and best practices outlined in the ACL Code of Ethics. Language Models can sometimes produce illogical or inaccurate reasoning paths, so their outputs should be cautiously used. The outputs are only examined to understand how a model arrives at its answers and investigate why it makes certain errors. All experiments used publicly available datasets from previously published works and did not involve ethical or privacy issues. 


\bibliography{acl2023}

\begin{thebibliography}{23}
\expandafter\ifx\csname natexlab\endcsname\relax\def\natexlab#1{#1}\fi

\bibitem[{Balakrishnan et~al.(2019)Balakrishnan, Rao, Upasani, White, and Subba}]{cdDialogue}
Anusha Balakrishnan, Jinfeng Rao, Kartikeya Upasani, Michael White, and Rajen Subba. 2019.
\newblock \href {https://doi.org/10.18653/V1/P19-1080} {Constrained decoding for neural {NLG} from compositional representations in task-oriented dialogue}.
\newblock In \emph{Proceedings of the 57th Conference of the Association for Computational Linguistics, {ACL} 2019, Florence, Italy, July 28- August 2, 2019, Volume 1: Long Papers}, pages 831--844. Association for Computational Linguistics.

\bibitem[{Besta et~al.(2024)Besta, Blach, Kubicek, Gerstenberger, Podstawski, Gianinazzi, Gajda, Lehmann, Niewiadomski, Nyczyk, and Hoefler}]{Got24}
Maciej Besta, Nils Blach, Ales Kubicek, Robert Gerstenberger, Michal Podstawski, Lukas Gianinazzi, Joanna Gajda, Tomasz Lehmann, Hubert Niewiadomski, Piotr Nyczyk, and Torsten Hoefler. 2024.
\newblock \href {https://doi.org/10.1609/AAAI.V38I16.29720} {Graph of thoughts: Solving elaborate problems with large language models}.
\newblock In \emph{Thirty-Eighth {AAAI} Conference on Artificial Intelligence, {AAAI} 2024, Thirty-Sixth Conference on Innovative Applications of Artificial Intelligence, {IAAI} 2024, Fourteenth Symposium on Educational Advances in Artificial Intelligence, {EAAI} 2014, February 20-27, 2024, Vancouver, Canada}, pages 17682--17690. {AAAI} Press.

\bibitem[{Brown et~al.(2020)Brown, Mann, Ryder, Subbiah, Kaplan, Dhariwal, Neelakantan, Shyam, Sastry, Askell, Agarwal, Herbert{-}Voss, Krueger, Henighan, Child, Ramesh, Ziegler, Wu, Winter, Hesse, Chen, Sigler, Litwin, Gray, Chess, Clark, Berner, McCandlish, Radford, Sutskever, and Amodei}]{gpt3}
Tom~B. Brown, Benjamin Mann, Nick Ryder, Melanie Subbiah, Jared Kaplan, Prafulla Dhariwal, Arvind Neelakantan, Pranav Shyam, Girish Sastry, Amanda Askell, Sandhini Agarwal, Ariel Herbert{-}Voss, Gretchen Krueger, Tom Henighan, Rewon Child, Aditya Ramesh, Daniel~M. Ziegler, Jeffrey Wu, Clemens Winter, Christopher Hesse, Mark Chen, Eric Sigler, Mateusz Litwin, Scott Gray, Benjamin Chess, Jack Clark, Christopher Berner, Sam McCandlish, Alec Radford, Ilya Sutskever, and Dario Amodei. 2020.
\newblock \href {http://arxiv.org/abs/2005.14165} {Language models are few-shot learners}.
\newblock \emph{CoRR}, abs/2005.14165.

\bibitem[{Chen et~al.(2023)Chen, Ma, Wang, and Cohen}]{Pot23}
Wenhu Chen, Xueguang Ma, Xinyi Wang, and William~W. Cohen. 2023.
\newblock \href {https://openreview.net/forum?id=YfZ4ZPt8zd} {Program of thoughts prompting: Disentangling computation from reasoning for numerical reasoning tasks}.
\newblock \emph{Transactions on Machine Learning Research}.

\bibitem[{Gou et~al.(2024)Gou, Shao, Gong, yelong shen, Yang, Duan, and Chen}]{critic}
Zhibin Gou, Zhihong Shao, Yeyun Gong, yelong shen, Yujiu Yang, Nan Duan, and Weizhu Chen. 2024.
\newblock \href {https://openreview.net/forum?id=Sx038qxjek} {{CRITIC}: Large language models can self-correct with tool-interactive critiquing}.
\newblock In \emph{The Twelfth International Conference on Learning Representations}.

\bibitem[{Hokamp and Liu(2017)}]{cdbeam}
Chris Hokamp and Qun Liu. 2017.
\newblock \href {https://doi.org/10.18653/V1/P17-1141} {Lexically constrained decoding for sequence generation using grid beam search}.
\newblock In \emph{Proceedings of the 55th Annual Meeting of the Association for Computational Linguistics, {ACL} 2017, Vancouver, Canada, July 30 - August 4, Volume 1: Long Papers}, pages 1535--1546. Association for Computational Linguistics.

\bibitem[{Lester et~al.(2020)Lester, Pressel, Hemmeter, Choudhury, and Bangalore}]{cdbeamNER}
Brian Lester, Daniel Pressel, Amy Hemmeter, Sagnik~Ray Choudhury, and Srinivas Bangalore. 2020.
\newblock \href {https://doi.org/10.18653/V1/2020.FINDINGS-EMNLP.166} {Constrained decoding for computationally efficient named entity recognition taggers}.
\newblock In \emph{Findings of the Association for Computational Linguistics: {EMNLP} 2020, Online Event, 16-20 November 2020}, volume {EMNLP} 2020 of \emph{Findings of {ACL}}, pages 1841--1848. Association for Computational Linguistics.

\bibitem[{Li et~al.(2022)Li, Lei, and Yang}]{Li2022FromEH}
Xin-Yi Li, Weixian Lei, and Yubin Yang. 2022.
\newblock \href {https://api.semanticscholar.org/CorpusID:249017531} {From easy to hard: Two-stage selector and reader for multi-hop question answering}.
\newblock \emph{ICASSP 2023 - 2023 IEEE International Conference on Acoustics, Speech and Signal Processing (ICASSP)}, pages 1--5.

\bibitem[{Och and Ney(2004)}]{beamori}
Franz~Josef Och and Hermann Ney. 2004.
\newblock \href {https://doi.org/10.1162/0891201042544884} {The alignment template approach to statistical machine translation}.
\newblock \emph{Comput. Linguistics}, 30(4):417--449.

\bibitem[{OpenAI et~al.(2024)OpenAI, Achiam, Adler, Agarwal, Ahmad, Akkaya, Aleman, Almeida, Altenschmidt, Altman, Anadkat, Avila, Babuschkin, Balaji, Balcom, Baltescu, Bao, Bavarian, Belgum, Bello, Berdine, Bernadett-Shapiro, Berner, Bogdonoff, Boiko, Boyd, Brakman, Brockman, Brooks, Brundage, Button, Cai, Campbell, Cann, Carey, Carlson, Carmichael, Chan, Chang, Chantzis, Chen, Chen, Chen, Chen, Chen, Chess, Cho, Chu, Chung, Cummings, Currier, Dai, Decareaux, Degry, Deutsch, Deville, Dhar, Dohan, Dowling, Dunning, Ecoffet, Eleti, Eloundou, Farhi, Fedus, Felix, Fishman, Forte, Fulford, Gao, Georges, Gibson, Goel, Gogineni, Goh, Gontijo-Lopes, Gordon, Grafstein, Gray, Greene, Gross, Gu, Guo, Hallacy, Han, Harris, He, Heaton, Heidecke, Hesse, Hickey, Hickey, Hoeschele, Houghton, Hsu, Hu, Hu, Huizinga, Jain, Jain, Jang, Jiang, Jiang, Jin, Jin, Jomoto, Jonn, Jun, Kaftan, Łukasz Kaiser, Kamali, Kanitscheider, Keskar, Khan, Kilpatrick, Kim, Kim, Kim, Kirchner, Kiros, Knight, Kokotajlo, Łukasz Kondraciuk,
  Kondrich, Konstantinidis, Kosic, Krueger, Kuo, Lampe, Lan, Lee, Leike, Leung, Levy, Li, Lim, Lin, Lin, Litwin, Lopez, Lowe, Lue, Makanju, Malfacini, Manning, Markov, Markovski, Martin, Mayer, Mayne, McGrew, McKinney, McLeavey, McMillan, McNeil, Medina, Mehta, Menick, Metz, Mishchenko, Mishkin, Monaco, Morikawa, Mossing, Mu, Murati, Murk, Mély, Nair, Nakano, Nayak, Neelakantan, Ngo, Noh, Ouyang, O'Keefe, Pachocki, Paino, Palermo, Pantuliano, Parascandolo, Parish, Parparita, Passos, Pavlov, Peng, Perelman, de~Avila Belbute~Peres, Petrov, de~Oliveira~Pinto, Michael, Pokorny, Pokrass, Pong, Powell, Power, Power, Proehl, Puri, Radford, Rae, Ramesh, Raymond, Real, Rimbach, Ross, Rotsted, Roussez, Ryder, Saltarelli, Sanders, Santurkar, Sastry, Schmidt, Schnurr, Schulman, Selsam, Sheppard, Sherbakov, Shieh, Shoker, Shyam, Sidor, Sigler, Simens, Sitkin, Slama, Sohl, Sokolowsky, Song, Staudacher, Such, Summers, Sutskever, Tang, Tezak, Thompson, Tillet, Tootoonchian, Tseng, Tuggle, Turley, Tworek, Uribe, Vallone,
  Vijayvergiya, Voss, Wainwright, Wang, Wang, Wang, Ward, Wei, Weinmann, Welihinda, Welinder, Weng, Weng, Wiethoff, Willner, Winter, Wolrich, Wong, Workman, Wu, Wu, Wu, Xiao, Xu, Yoo, Yu, Yuan, Zaremba, Zellers, Zhang, Zhang, Zhao, Zheng, Zhuang, Zhuk, and Zoph}]{openai2024gpt4}
OpenAI, Josh Achiam, Steven Adler, Sandhini Agarwal, Lama Ahmad, Ilge Akkaya, Florencia~Leoni Aleman, Diogo Almeida, Janko Altenschmidt, Sam Altman, Shyamal Anadkat, Red Avila, Igor Babuschkin, Suchir Balaji, Valerie Balcom, Paul Baltescu, Haiming Bao, Mohammad Bavarian, Jeff Belgum, Irwan Bello, Jake Berdine, Gabriel Bernadett-Shapiro, Christopher Berner, Lenny Bogdonoff, Oleg Boiko, Madelaine Boyd, Anna-Luisa Brakman, Greg Brockman, Tim Brooks, Miles Brundage, Kevin Button, Trevor Cai, Rosie Campbell, Andrew Cann, Brittany Carey, Chelsea Carlson, Rory Carmichael, Brooke Chan, Che Chang, Fotis Chantzis, Derek Chen, Sully Chen, Ruby Chen, Jason Chen, Mark Chen, Ben Chess, Chester Cho, Casey Chu, Hyung~Won Chung, Dave Cummings, Jeremiah Currier, Yunxing Dai, Cory Decareaux, Thomas Degry, Noah Deutsch, Damien Deville, Arka Dhar, David Dohan, Steve Dowling, Sheila Dunning, Adrien Ecoffet, Atty Eleti, Tyna Eloundou, David Farhi, Liam Fedus, Niko Felix, Simón~Posada Fishman, Juston Forte, Isabella Fulford, Leo
  Gao, Elie Georges, Christian Gibson, Vik Goel, Tarun Gogineni, Gabriel Goh, Rapha Gontijo-Lopes, Jonathan Gordon, Morgan Grafstein, Scott Gray, Ryan Greene, Joshua Gross, Shixiang~Shane Gu, Yufei Guo, Chris Hallacy, Jesse Han, Jeff Harris, Yuchen He, Mike Heaton, Johannes Heidecke, Chris Hesse, Alan Hickey, Wade Hickey, Peter Hoeschele, Brandon Houghton, Kenny Hsu, Shengli Hu, Xin Hu, Joost Huizinga, Shantanu Jain, Shawn Jain, Joanne Jang, Angela Jiang, Roger Jiang, Haozhun Jin, Denny Jin, Shino Jomoto, Billie Jonn, Heewoo Jun, Tomer Kaftan, Łukasz Kaiser, Ali Kamali, Ingmar Kanitscheider, Nitish~Shirish Keskar, Tabarak Khan, Logan Kilpatrick, Jong~Wook Kim, Christina Kim, Yongjik Kim, Jan~Hendrik Kirchner, Jamie Kiros, Matt Knight, Daniel Kokotajlo, Łukasz Kondraciuk, Andrew Kondrich, Aris Konstantinidis, Kyle Kosic, Gretchen Krueger, Vishal Kuo, Michael Lampe, Ikai Lan, Teddy Lee, Jan Leike, Jade Leung, Daniel Levy, Chak~Ming Li, Rachel Lim, Molly Lin, Stephanie Lin, Mateusz Litwin, Theresa Lopez, Ryan
  Lowe, Patricia Lue, Anna Makanju, Kim Malfacini, Sam Manning, Todor Markov, Yaniv Markovski, Bianca Martin, Katie Mayer, Andrew Mayne, Bob McGrew, Scott~Mayer McKinney, Christine McLeavey, Paul McMillan, Jake McNeil, David Medina, Aalok Mehta, Jacob Menick, Luke Metz, Andrey Mishchenko, Pamela Mishkin, Vinnie Monaco, Evan Morikawa, Daniel Mossing, Tong Mu, Mira Murati, Oleg Murk, David Mély, Ashvin Nair, Reiichiro Nakano, Rajeev Nayak, Arvind Neelakantan, Richard Ngo, Hyeonwoo Noh, Long Ouyang, Cullen O'Keefe, Jakub Pachocki, Alex Paino, Joe Palermo, Ashley Pantuliano, Giambattista Parascandolo, Joel Parish, Emy Parparita, Alex Passos, Mikhail Pavlov, Andrew Peng, Adam Perelman, Filipe de~Avila Belbute~Peres, Michael Petrov, Henrique~Ponde de~Oliveira~Pinto, Michael, Pokorny, Michelle Pokrass, Vitchyr~H. Pong, Tolly Powell, Alethea Power, Boris Power, Elizabeth Proehl, Raul Puri, Alec Radford, Jack Rae, Aditya Ramesh, Cameron Raymond, Francis Real, Kendra Rimbach, Carl Ross, Bob Rotsted, Henri Roussez,
  Nick Ryder, Mario Saltarelli, Ted Sanders, Shibani Santurkar, Girish Sastry, Heather Schmidt, David Schnurr, John Schulman, Daniel Selsam, Kyla Sheppard, Toki Sherbakov, Jessica Shieh, Sarah Shoker, Pranav Shyam, Szymon Sidor, Eric Sigler, Maddie Simens, Jordan Sitkin, Katarina Slama, Ian Sohl, Benjamin Sokolowsky, Yang Song, Natalie Staudacher, Felipe~Petroski Such, Natalie Summers, Ilya Sutskever, Jie Tang, Nikolas Tezak, Madeleine~B. Thompson, Phil Tillet, Amin Tootoonchian, Elizabeth Tseng, Preston Tuggle, Nick Turley, Jerry Tworek, Juan Felipe~Cerón Uribe, Andrea Vallone, Arun Vijayvergiya, Chelsea Voss, Carroll Wainwright, Justin~Jay Wang, Alvin Wang, Ben Wang, Jonathan Ward, Jason Wei, CJ~Weinmann, Akila Welihinda, Peter Welinder, Jiayi Weng, Lilian Weng, Matt Wiethoff, Dave Willner, Clemens Winter, Samuel Wolrich, Hannah Wong, Lauren Workman, Sherwin Wu, Jeff Wu, Michael Wu, Kai Xiao, Tao Xu, Sarah Yoo, Kevin Yu, Qiming Yuan, Wojciech Zaremba, Rowan Zellers, Chong Zhang, Marvin Zhang, Shengjia
  Zhao, Tianhao Zheng, Juntang Zhuang, William Zhuk, and Barret Zoph. 2024.
\newblock \href {http://arxiv.org/abs/2303.08774} {Gpt-4 technical report}.

\bibitem[{Post and Vilar(2018)}]{cdbeam2}
Matt Post and David Vilar. 2018.
\newblock \href {https://doi.org/10.18653/V1/N18-1119} {Fast lexically constrained decoding with dynamic beam allocation for neural machine translation}.
\newblock In \emph{Proceedings of the 2018 Conference of the North American Chapter of the Association for Computational Linguistics: Human Language Technologies, {NAACL-HLT} 2018, New Orleans, Louisiana, USA, June 1-6, 2018, Volume 1 (Long Papers)}, pages 1314--1324. Association for Computational Linguistics.

\bibitem[{Sel et~al.(2024)Sel, Tawaha, Khattar, Jia, and Jin}]{Aot24}
Bilgehan Sel, Ahmad Tawaha, Vanshaj Khattar, Ruoxi Jia, and Ming Jin. 2024.
\newblock \href {https://openreview.net/forum?id=KJL2b6BthC} {Algorithm of thoughts: Enhancing exploration of ideas in large language models}.
\newblock In \emph{Forty-first International Conference on Machine Learning}.

\bibitem[{Touvron et~al.(2023)Touvron, Martin, Stone, Albert, Almahairi, Babaei, Bashlykov, Batra, Bhargava, Bhosale, Bikel, Blecher, Ferrer, Chen, Cucurull, Esiobu, Fernandes, Fu, Fu, Fuller, Gao, Goswami, Goyal, Hartshorn, Hosseini, Hou, Inan, Kardas, Kerkez, Khabsa, Kloumann, Korenev, Koura, Lachaux, Lavril, Lee, Liskovich, Lu, Mao, Martinet, Mihaylov, Mishra, Molybog, Nie, Poulton, Reizenstein, Rungta, Saladi, Schelten, Silva, Smith, Subramanian, Tan, Tang, Taylor, Williams, Kuan, Xu, Yan, Zarov, Zhang, Fan, Kambadur, Narang, Rodriguez, Stojnic, Edunov, and Scialom}]{llama2}
Hugo Touvron, Louis Martin, Kevin Stone, Peter Albert, Amjad Almahairi, Yasmine Babaei, Nikolay Bashlykov, Soumya Batra, Prajjwal Bhargava, Shruti Bhosale, Dan Bikel, Lukas Blecher, Cristian~Canton Ferrer, Moya Chen, Guillem Cucurull, David Esiobu, Jude Fernandes, Jeremy Fu, Wenyin Fu, Brian Fuller, Cynthia Gao, Vedanuj Goswami, Naman Goyal, Anthony Hartshorn, Saghar Hosseini, Rui Hou, Hakan Inan, Marcin Kardas, Viktor Kerkez, Madian Khabsa, Isabel Kloumann, Artem Korenev, Punit~Singh Koura, Marie-Anne Lachaux, Thibaut Lavril, Jenya Lee, Diana Liskovich, Yinghai Lu, Yuning Mao, Xavier Martinet, Todor Mihaylov, Pushkar Mishra, Igor Molybog, Yixin Nie, Andrew Poulton, Jeremy Reizenstein, Rashi Rungta, Kalyan Saladi, Alan Schelten, Ruan Silva, Eric~Michael Smith, Ranjan Subramanian, Xiaoqing~Ellen Tan, Binh Tang, Ross Taylor, Adina Williams, Jian~Xiang Kuan, Puxin Xu, Zheng Yan, Iliyan Zarov, Yuchen Zhang, Angela Fan, Melanie Kambadur, Sharan Narang, Aurelien Rodriguez, Robert Stojnic, Sergey Edunov, and Thomas
  Scialom. 2023.
\newblock \href {http://arxiv.org/abs/2307.09288} {Llama 2: Open foundation and fine-tuned chat models}.

\bibitem[{Trivedi et~al.(2022)Trivedi, Balasubramanian, Khot, and Sabharwal}]{TrivediBKS22}
Harsh Trivedi, Niranjan Balasubramanian, Tushar Khot, and Ashish Sabharwal. 2022.
\newblock \href {https://doi.org/10.1162/TACL\_A\_00475} {Musique: Multihop questions via single-hop question composition}.
\newblock \emph{Trans. Assoc. Comput. Linguistics}, 10:539--554.

\bibitem[{Wang et~al.(2023)Wang, Wei, Schuurmans, Le, Chi, Narang, Chowdhery, and Zhou}]{sccot}
Xuezhi Wang, Jason Wei, Dale Schuurmans, Quoc~V. Le, Ed~H. Chi, Sharan Narang, Aakanksha Chowdhery, and Denny Zhou. 2023.
\newblock \href {https://openreview.net/pdf?id=1PL1NIMMrw} {Self-consistency improves chain of thought reasoning in language models}.
\newblock In \emph{The Eleventh International Conference on Learning Representations, {ICLR} 2023, Kigali, Rwanda, May 1-5, 2023}. OpenReview.net.

\bibitem[{Wei et~al.(2022)Wei, Wang, Schuurmans, Bosma, Ichter, Xia, Chi, Le, and Zhou}]{CotWei}
Jason Wei, Xuezhi Wang, Dale Schuurmans, Maarten Bosma, Brian Ichter, Fei Xia, Ed~H. Chi, Quoc~V. Le, and Denny Zhou. 2022.
\newblock \href {http://papers.nips.cc/paper\_files/paper/2022/hash/9d5609613524ecf4f15af0f7b31abca4-Abstract-Conference.html} {Chain-of-thought prompting elicits reasoning in large language models}.
\newblock In \emph{Advances in Neural Information Processing Systems 35: Annual Conference on Neural Information Processing Systems 2022, NeurIPS 2022, New Orleans, LA, USA, November 28 - December 9, 2022}.

\bibitem[{Welbl et~al.(2018)Welbl, Stenetorp, and Riedel}]{WelblSR18}
Johannes Welbl, Pontus Stenetorp, and Sebastian Riedel. 2018.
\newblock \href {https://doi.org/10.1162/TACL\_A\_00021} {Constructing datasets for multi-hop reading comprehension across documents}.
\newblock \emph{Trans. Assoc. Comput. Linguistics}, 6:287--302.

\bibitem[{Yang et~al.(2018)Yang, Qi, Zhang, Bengio, Cohen, Salakhutdinov, and Manning}]{yang2018hotpotqa}
Zhilin Yang, Peng Qi, Saizheng Zhang, Yoshua Bengio, William~W. Cohen, Ruslan Salakhutdinov, and Christopher~D. Manning. 2018.
\newblock {HotpotQA}: A dataset for diverse, explainable multi-hop question answering.
\newblock In \emph{Conference on Empirical Methods in Natural Language Processing ({EMNLP})}.

\bibitem[{Yao et~al.(2023{\natexlab{a}})Yao, Yu, Zhao, Shafran, Griffiths, Cao, and Narasimhan}]{TOT23}
Shunyu Yao, Dian Yu, Jeffrey Zhao, Izhak Shafran, Tom Griffiths, Yuan Cao, and Karthik Narasimhan. 2023{\natexlab{a}}.
\newblock \href {http://papers.nips.cc/paper\_files/paper/2023/hash/271db9922b8d1f4dd7aaef84ed5ac703-Abstract-Conference.html} {Tree of thoughts: Deliberate problem solving with large language models}.
\newblock In \emph{Advances in Neural Information Processing Systems 36: Annual Conference on Neural Information Processing Systems 2023, NeurIPS 2023, New Orleans, LA, USA, December 10 - 16, 2023}.

\bibitem[{Yao et~al.(2023{\natexlab{b}})Yao, Zhao, Yu, Du, Shafran, Narasimhan, and Cao}]{react}
Shunyu Yao, Jeffrey Zhao, Dian Yu, Nan Du, Izhak Shafran, Karthik~R. Narasimhan, and Yuan Cao. 2023{\natexlab{b}}.
\newblock \href {https://openreview.net/pdf?id=WE\_vluYUL-X} {React: Synergizing reasoning and acting in language models}.
\newblock In \emph{The Eleventh International Conference on Learning Representations, {ICLR} 2023, Kigali, Rwanda, May 1-5, 2023}. OpenReview.net.

\bibitem[{Yin et~al.(2023)Yin, Wang, Hu, Wu, Yan, Zhang, Cao, Huang, and Qiu}]{YinWHWYZCHQ23}
Zhangyue Yin, Yuxin Wang, Xiannian Hu, Yiguang Wu, Hang Yan, Xinyu Zhang, Zhao Cao, Xuanjing Huang, and Xipeng Qiu. 2023.
\newblock \href {https://doi.org/10.1007/978-981-99-6207-5\_5} {Rethinking label smoothing on multi-hop question answering}.
\newblock In \emph{Chinese Computational Linguistics - 22nd China National Conference, {CCL} 2023, Harbin, China, August 3-5, 2023, Proceedings}, volume 14232 of \emph{Lecture Notes in Computer Science}, pages 72--87. Springer.

\bibitem[{Zhang et~al.(2023)Zhang, Zhang, Zhang, Liu, and Huang}]{beam}
Jiahao Zhang, Haiyang Zhang, Dongmei Zhang, Yong Liu, and Shen Huang. 2023.
\newblock \href {https://doi.org/10.48550/ARXIV.2308.08973} {Beam retrieval: General end-to-end retrieval for multi-hop question answering}.
\newblock \emph{CoRR}, abs/2308.08973.

\bibitem[{Zhu et~al.(2021)Zhu, Lei, Wang, Zheng, Poria, and Chua}]{zhu2021}
Fengbin Zhu, Wenqiang Lei, Chao Wang, Jianming Zheng, Soujanya Poria, and Tat{-}Seng Chua. 2021.
\newblock \href {http://arxiv.org/abs/2101.00774} {Retrieving and reading: {A} comprehensive survey on open-domain question answering}.
\newblock \emph{CoRR}, abs/2101.00774.

\end{thebibliography}
\bibliographystyle{acl_natbib}

\clearpage
\appendix

\section{Prompt Templates}
\label{sec:appendixA}
\paragraph{Sub Question Generation Template} 
The prompt template containing one comparison question and one bridge question is given below:

\textbf{prompt}: Break a question into high-quality sub-questions that are easier to answer. Here are two examples as guidelines:\\
"Question: Are Tokyo and Busan in the same country? Thought 1: I could either find which country Tokyo is located in, or which country Busan is located in. Sub Question 1-1: Which country is Tokyo located in? Sub Question 1-2: Which country is Busan located in?"\\
"Question: Tokyo is located in the country that has what colors present on its national flag? Thought 1: I need to first find out which country Tokyo is located in. Sub Question 1-1: Which country is Tokyo located in?" \\
Only give out your thought process and current-level sub-questions. Do not give out answers to your questions. Question: \textit{Given Question}. Thought 1:

\paragraph{Prompt-based Constrained Generation Template}
The prompt template at answering time is given below:

\textbf{prompt}:
Given a question and a list of evidence that may of help, give your answer directly, using words only from the vocabulary bank, without any explanations. \\
Question: \textit{Given Question}. Evidence as reference: \textit{Given Evidence}. Vocabulary Bank: \textit{Given Vocabulary}. Answer:

\section{Examples of the Error Cases}
\label{sec:appendixB}

\textbf{\textbullet{Type-2: Intermediate Answer}}\\
\rule{\linewidth}{0.4pt}
\textit{Question}:\\
Where does the hotel and casino located in which Bill Cosby's third album was recorded?\\
\textit{Answer} given by \our on GPT4:\\
Las Vegas.\\
\textit{Golden Answer}:\\
Las Vegas Strip in Paradise.\\

\textbf{\textbullet{Type-3: Wrong Answer}}\\
\rule{\linewidth}{0.4pt}
\textit{Question}:\\
Aside from the Apple Remote, what other device can control the program Apple Remote was originally designed to interact with?\\
\textit{Answer} given by \our on GPT4:\\
siri remote and devices with netsupport manager software\\
\textit{Golden Answer}:\\
keyboard function keys\\

\textbf{\textbullet{Type-4: Semantically Correct}}\\
\rule{\linewidth}{0.4pt}
\textit{Question}:\\
Roger O. Egeberg was Assistant Secretary for Health and Scientific Affairs during the administration of a president that served during what years?\\
\textit{Answer} given by \our on GPT4:\\
1969 to 1974\\
\textit{Golden Answer}:\\
1969 until 1974\\

\end{document}